# Assembly and Disassembly Planning by using Fuzzy Logic & Genetic Algorithms

**L . M. Galantucci; G. Percoco & R. Spina**
Dipartimento di Ingegneria Meccanica e Gestionale (DIMeG)
Politecnico di Bari, Viale Japigia, 182
70126 Bari, Italy

***Abstract:*** *The authors propose the implementation of hybrid Fuzzy Logic-Genetic Algorithm (FL-GA) methodology to plan the automatic assembly and disassembly sequence of products. The GA-Fuzzy Logic approach is implemented onto two levels. The first level of hybridization consists of the development of a Fuzzy controller for the parameters of an assembly or disassembly planner based on GAs. This controller acts on mutation probability and crossover rate in order to adapt their values dynamically while the algorithm runs. The second level consists of the identification of the optimal assembly or disassembly sequence by a Fuzzy function, in order to obtain a closer control of the technological knowledge of the assembly/disassembly process. Two case studies were analyzed in order to test the efficiency of the Fuzzy-GA methodologies.*
***Keywords:*** *Genetic Algorithms, Fuzzy Logic, Assembly planning, Disassembly planning*

## 1. Introduction

The rapid development of new products has shortened product time-to-market and shelf-life, increasing the quantity of wasted used goods. In this context, all these factors must be considered during the design stage of a product. Design for X (DFX) is a collection of methodologies which allow the correct evaluation of several product characteristics and requirements at the earliest stage of the development cycle by enclosing several design aspects (i.e. assembly, disassembly, manufacturability, etc.). Assembly and disassembly are two processes that receive a lot of benefits from DFX. The assembly process is one of the most time-consuming and expensive manufacturing activities. Disassembly aspects have to be taken into account in several steps of the product Life Cycle, both in product design and during the process design for the disassembly of End-of-Life products. The main objectives of disassembly are the maintenance, remanufacturing, recycling or disposal of end-of-life products. As the complexity of products and production systems increases, the need for computer mediated design tools that aid designers in dealing with assembly and disassembly aspects is becoming greater (Boothroyd G., & Alting L., 1992). The development of efficient algorithms and computer aided integrated methods to evaluate the effectiveness of assembly and disassembly sequences is necessary. Efficiency and flexibility to operate with the maximum number of different products, production environment and plant-layouts are the main features of these algorithms.

In this paper the authors propose a hybrid Fuzzy Logic–Genetic Algorithm approach to implement the automatic generation of optimal assembly and disassembly sequences. The aim of this methodology is the efficient generation of these sequences while preserving the flexibility to operate with a great variety of industrial products and assembly/disassembly environments.

## 2. Research Background

The realization of an efficient assembly/disassembly (A/D) process is a key factor for the competitiveness of successful company. Environmental regulations and customer pressure to make more environmentally friendly products, force companies to further integrate A/D into the manufacturing environment. However, the efficiency of existing A/D processes is still low because of the inherent difficulty to develop fully automated systems and the complexity of performing operations for a wide range of products (Guide V.D.R., 2000).
Several methodologies have been proposed by academic and industrial researchers in order to implement automated A/D systems. These methodologies fall into three main research areas related to: (i) the conception of flexible A/D cells, (ii) design and develop of innovative A/D equipments and tools, (iii) implementation of more efficient control systems and (iv) formalization of predictive models for A/D planning. The implementation of a fully automated A/D cell requires a medium-high capital investment. The main reason for automating A/D cells is their flexibility to cope with a great number of factors such as: product variety, small batch sizes, different product states, missing components and sometimes inadequate disassembly tools. The cell can also work with new equipment and tools. In this way the disassembly cycle can be shortened using special devices for separating, unscrewing and grasping (Santochi M., Dini G., Failli F., 2002). In addition specialized active sensors and/or devices must be studied and realized to assure the necessary reliability and speed during on-line product inspection because of the limited time for data acquisition and elaboration. For this reason, a stereoscopic vision system is considered more efficient than others for recognizing product features (Buker U. et alii, 2001). Another important aspect is the integration between the several hardware elements of the disassembly cells (robots, tool changing station, etc.) and disassembly planning software. Such software, normally tailor-made and based on the specific application (e.g. mechanical or electronic parts), require the formalization of an extensive process-oriented knowledge. This knowledge can be embedded into the system through procedures applied to the verification model obtained from the CAD system in order to make comparisons with the real product. The inspection system verifies the state of maintenance of the product. Recognized differences between the product model and physical part are then transferred to the planner and the generation of new optimal sequences is carried out (Salomonski N., & Zussman E. 1999).

*2.1. Assembly and Disassembly Planning*

One of the most challenging aspects of A/D is finding the optimal sequence. The assembly sequence is traditionally generated by a human expert who carefully studies the assembly drawing and generates the sequence in his mind. This planning step is very costly and time consuming. Together with time and cost issues, manufacturers are becoming more environmentally sensible. In addition, stricter regulations are forcing manufacturers to become more responsible for the entire product life cycle. As a consequence, the evaluation of disassembly and recycling of products has to be considered at the design stage, planning all the operations to be performed on end-of-life products and taking into consideration different part conditions because of deterioration (i.e. damaged or missing parts).

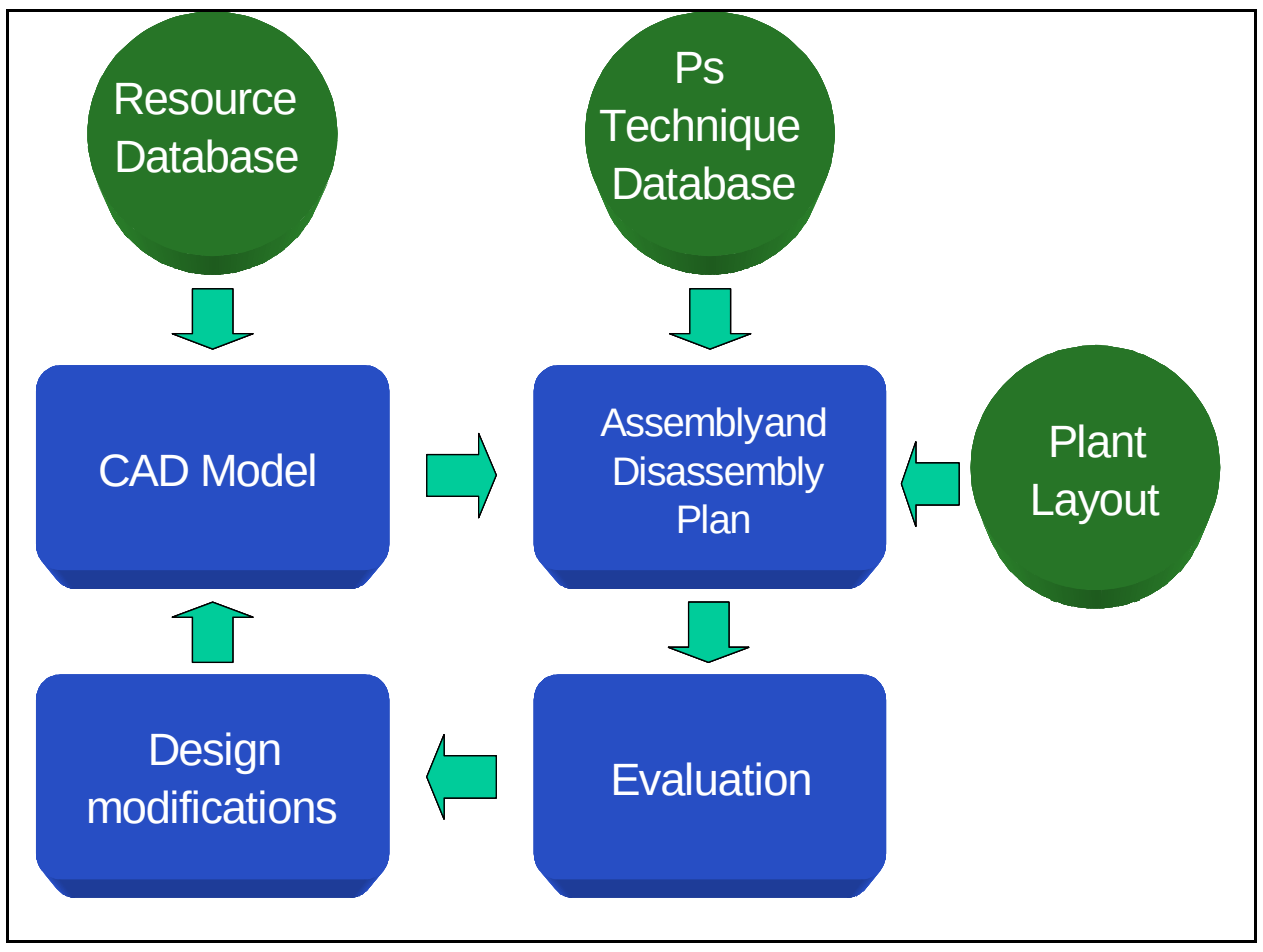

Fig. 1. CAD based Assembly/Disassembly system

In the process of disposing and recycling end-of-life products, the cost of handling, sorting and disassembly will play an ever more important role.
The aim of assembly planning is the identification of the optimal assembly sequences of products constituted of several parts, whereas disassembly planning is very important during product maintenance or end-of-life applications. In particular, the disassembly sequence necessary to realize efficient product maintenance can be identified by inverting the assembly sequence. This condition is necessary to preserve the product functionality, avoiding destructive operations on components that can lead to useless products. In end-of-life applications, the main aspects influencing the generation of disassembly sequences are the evaluation of the financial and environmental impact of waste (analysis of materials, energy and toxicity is necessary in case of remanufacturing, recycling or disposal) and economic aspects associated with the feasibility of the disassembly process (Harjula T., & al., 1996).
The development of a system for the automatic recognition of assembly and disassembly features for the generation of sequences is currently in a research phase. In such a system (Srinivasan H. et al., 1999), data on assembly and disassembly features and related resources are retrieved from the CAD and Resource Database in order to plan operation sequences (Fig. 1). The role of the CAD system is essentially connected to data collection and analysis of the assembly/disassembly process. During data collection, CAD models of the different components are re-organized into one model in order to restore the original configuration of the product. Assembly and Disassembly delations (contact, attachment and blocking between components) and directions are identified. Related resources (grippers and fixtures) linked to disassembly features are then selected to prepare the planning phase. The creation of these sequences, checks on their feasibility and the identification of the optimal sequence are the principal tasks performed by the assembly and disassembly planner. The planner must take into consideration the plant layout for the machine positions. In fact, the machine characteristics and resources influence the shape of the solution of the process (e.g. a change of grippers may have higher costs than re-orientation if performed on one machine, but it may be cheaper if performed on another one). The generated assembly and/or disassembly sequence is then evaluated by designers and process engineers. If an unsatisfactory sequence has been obtained, modifications are performed to the original model and/or to the problem solving technique in order to identify a more effective sequence plan. The successful application of these tasks is strictly dependent on the problem solving techniques used (PS Technique Database). The application of a well-developed problem-solving technique allows the reduction of the computational effort without affecting the reliability of the results of the processed model. Moreover, an objective criterion must be used to evaluate disassembly sequences. This function is normally built using some parameters such as the number of product orientations or the number and type of grippers used (Dini G. & Santochi M., 1992). Additional constraints and precedence between components can be added to deal with the real industrial environment. Furthermore, easy adaptability to a wide variety of industrial situations is required for the analysis of several products and machines.

*2.2. Formulation of the Disassembly Problem*

An A/D plan to perform the complete product union/separation is represented using well-specified sets, defined based on product features and manufacturing environment data (Subramani A.K., & Dewhurst P. 1991). These sets generally contain information on the position of each basic component in the A/D sequence (sequence set), tools used to connect/disconnect interfaces between the selected component and other elements (tool set) and the product orientation chosen during the connection/disconnection tasks (orientation set) (Kroll E., Carver B.S. 1999),. The problem of identifying the optimal A/D sequence presents a factorial computational complexity in function of the number of components n. This complexity rapidly grows up when the aim is the identification of the optimal A/D plan rather than the sequence. The complexity of A/D planning also becomes dependent on the number of grippers g and product orientations o. The ranking of the product components in the A/D plan is performed using a properly-designed objective function. This function is built by considering technological aspects, environmental considerations, A/D times and costs, or a combination of them. In addition the proposed solutions must respect constraints related to interactions between components during disassembly, possible grouping of components and directional constraints imposed on the component geometry. In literature, several methods for the problem formulation of A/D planning can be found. This formulation generally consists of two steps. The first one is the definition of the geometrical relationships and mating conditions between components.

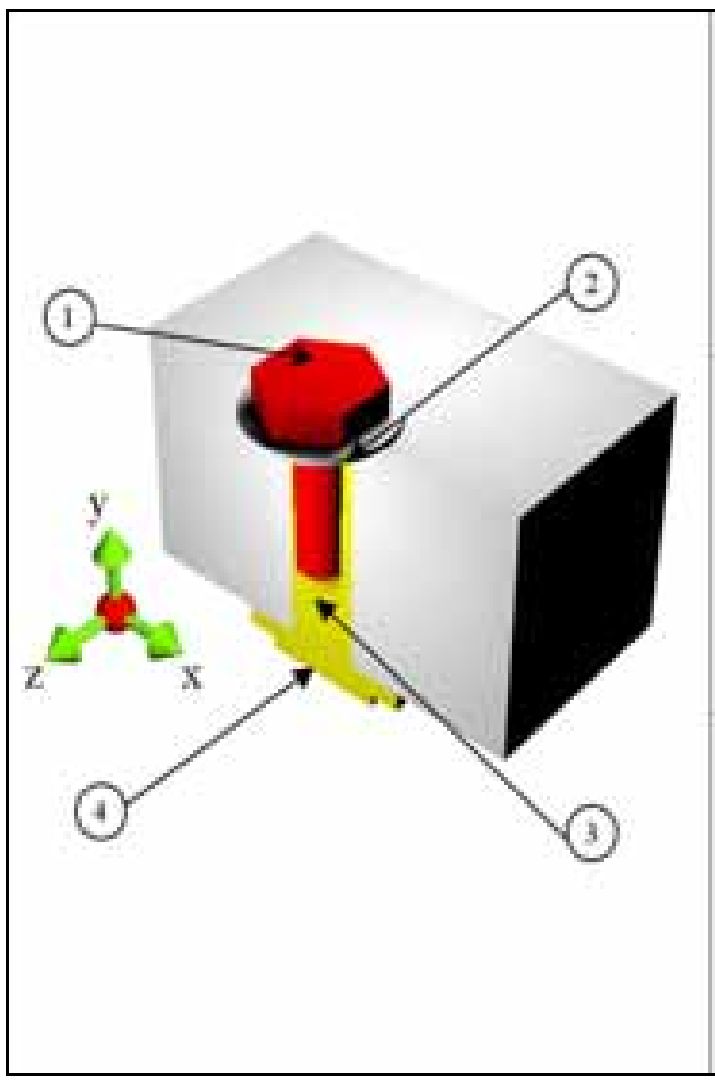

$A_y$

| | 1 | 2 | 3 | 4 |
|---|---|---|---|---|
| 1 | 0 | 0 | 0 | 0 |
| 2 | 1 | 0 | 0 | 0 |
| 3 | 0 | 1 | 0 | 0 |
| 4 | 1 | 1 | 0 | 0 |

$B_y$

| | 1 | 2 | 3 | 4 |
|---|---|---|---|---|
| 1 | 0 | 0 | 0 | 0 |
| 2 | 1 | 0 | 0 | 0 |
| 3 | 0 | 1 | 0 | 0 |
| 4 | 0 | 1 | 0 | 0 |

$C_y$

| | 1 | 2 | 3 | 4 |
|---|---|---|---|---|
| 1 | 0 | 0 | 0 | 1 |
| 2 | 0 | 0 | 0 | 0 |
| 3 | 0 | 0 | 0 | 0 |
| 4 | -1 | 1 | 0 | 0 |

Fig. 2. CAD based Assembly/Disassembly system

The second one is the formalization of the optimization method based on computation, technological and environmental aspects. With regard to the first step, the approaches developed in literature are basically graph-based and matrix-based methods. These approaches are equivalent because matrices can be built from graphs and vice versa. In graph-based approaches, relationships between components are represented using the graph semantic (Kaebernick H., et al., 2000) (Subramani A.K., & Dewhurst P. 1991), (Laperrière L., Elmaraghy H.A. 1992); in matrix-based approaches, relations are converted into matrix form (Dini G. & Santochi M., 1992) (Lazzerini B. & Marcelloni F., 2000). These approaches are very useful if a CAD-based system is to be employed. Main geometry and feature data are extracted from the CAD model and converted into matrix/graph form. A typical application of matrix methods is shown in Fig. 2, where three matrices (Ay, By, Cy) are defined for the considered assembly. The y specifies the considered A/D direction for each matrix. The Ay matrix is called interference matrix, its rows and columns correspond to each component and its elements can be only 0 or 1. Considering the ith row the ones identify the elements that do not permit the disassembly of the ith component in the y direction, the zeros are related to components which do not prevent it. The By matrix detects the components in contact with the ith component, while the Cy matrix identifies which components are mechanically connected to the ith component. Then the matrices and the gripper list are then sent to the planner together with the interference matrix. In a computer aided environment this information can be automatically retrieved from the CAD model. Once a component is removed, the row and column associated to it are removed. This process is repeated until only one component is left. The final output is represented by the sequence which optimizes the defined objective function. The second step is the formalization of the optimization method for the generic objective function:

$$f = f\,(C,T,E) \qquad (1)$$

where C, T and E are computation, technological and environmental factors respectively. Computation factors, basically associated to the maximum length of the feasible sequence, are introduced into the model to perform computations. Technological factors are related to the number of assembly or disassembly operations (gripper changes, product reorientations, fixture changes, etc.). Environmental factors are mainly related to subassembly detection, the theoretical value of the subassembly, weight of the subassembly and percentage of each material present in the subassembly (Dini G. & Failli F. 1998). For a product consisting of a small number of components, the optimal A/D plan can be found using an enumerative algorithm that generates and evaluates all component configurations. This approach becomes unfeasible with a high number of components because of the huge number of combinations. In fact, the factorial complexity grows faster than the polynomial one. Because of the combinational problem complexity, the size of the problem and the flexibility, required to solve the algorithm, influence the choice of the solution approach. An algorithm that works well on small size problems can become impossible to be applied to large problems. On the contrary too complex algorithms can be time and resource consuming. These circumstances lead to several approaches to model and solve the disassembly problem, also to limit the computational effort. The identification of the optimal disassembly plan of a product and/or the creation of viable A/D sequences are commonly performed using off-line methodologies such as: (i) operational research approaches, (ii) genetic algorithms, (iii) ant colony, (iv) simulated annealing, (v) Petri nets or (vi) a combination of the above (O'Shea B., et al. 1998). Genetic Algorithms (GAs) are particularly efficient in searching for optimal solutions. GAs are more robust than existing direct search methods (hill-climbing, simulated annealing, etc) because they present a multi-directional search in the solution space (set of individuals) and encourage information exchange between these directions (Michalewicz Z., 1999). GAs unfortunately show some drawbacks in this application, such as difficulties in finding the optimal value of genetic parameters, problems related to the search of optimal values of the weighted fitness function and difficulties related to multi-objective problems. In order to improve the GA behavior to avoid these drawbacks, an integrated approach with Fuzzy Logic has been proposed (see below).

## 3. Proposed approach and case studies

Actually there is an increasing interest in the integration of Fuzzy Logic (FL) and Genetic Algorithms (Gas). According to an exhaustive literature analysis (Cordón O., et al., 1997), this integration can be realized on two levels: (i) FL can be used to improve GA behavior and modeling GA components, resulting in fuzzy genetic algorithms (FGAs); (ii) the application of GAs in various optimization and search problems involving fuzzy systems (Genetic Fuzzy Systems – GFS).

A Fuzzy Genetic Algorithm (FGA) is a GA in which some algorithm components are implemented using fuzzy logic based tools, such as: fuzzy operators and fuzzy connectives for designing genetic operators with different properties, fuzzy logic control systems for controlling the GA parameters according to some performance measures, fuzzy stop criteria, etc.

FGAs are a particular kind of adaptive genetic algorithms (AGAs), characterized by a variation of their features during running based on some specific performance measures. A generic AGA can perform the adaptive setting of parameter values, the adaptive selection of genetic operators and produce an adaptive representation of the fitness function. Among these ways of making a GA adaptive, the adaptation of parameter setting has been widely studied. In particular, rules for discrete Fuzzy Logic Controllers of mutation probability and selection pressure have been formalized (cross-over probability, cross-over rate etc.) (Herrera F., & Lozano M., 1996). According to these rules, a high mutation probability and a low selection pressure must be used during the initial generation of individuals. The second integration method involves techniques such as Fuzzy clustering, Fuzzy optimization (GAs are used for solving different fuzzy optimization problems), Fuzzy neural networks (GAs are used for determining an optimal set of link weight and for participating in hybrid learning algorithms) and Fuzzy expert systems (GAs can solve basic problems of the knowledge base).

In this paper the authors propose a hybrid Fuzzy-GA methodology to identify the optimal assembly and disassembly sequences while improving performances of GAs. Two case studies have been selected (Dini G., et al. 1999) (Dini G., & Failli F. 1999) in order to investigate and validate the proposed approach. The identification of the optimal disassembly sequences of these products was carried using different methodologies. Fig. 3 and Fig. 4 report product configurations, the component list and available grippers to perform A/D operations.

| Part ID | Available Grippers |
|---|---|
| 1 | G1 G2 G4 |
| 2 | G1 G2 G4 |
| 3 | G1 G2 G3 |
| 4 | G3 |
| 5 | G1 G2 G4 |
| 6 | G1 G2 G4 |
| 7 | G1 G2 G4 |
| 8 | G3 |

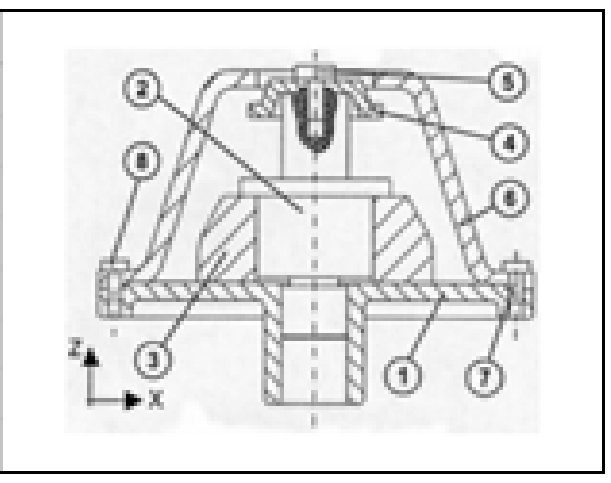

Fig. 3. First Product

| Part ID | Available Grippers |
|---|---|
| 1 | G1 G2 |
| 2 | G3 |
| 3 | G6 |
| 4 | G1 G2 G4 |
| 5 | G1 G3 |
| 6 | G1 G4 |
| 7 | G1 G2 |
| 8 | G3 G4 |
| 9 | G3 |
| 10 | G5 |
| 11 | G1 G2 G4 |
| 12 | G3 |
| 13 | G3 |
| 14 | G3 G4 |
| 15 | G7 |
| 16 | G7 |
| 17 | G2 |
| 18 | G1 G2 G5 |
| 19 | G8 |

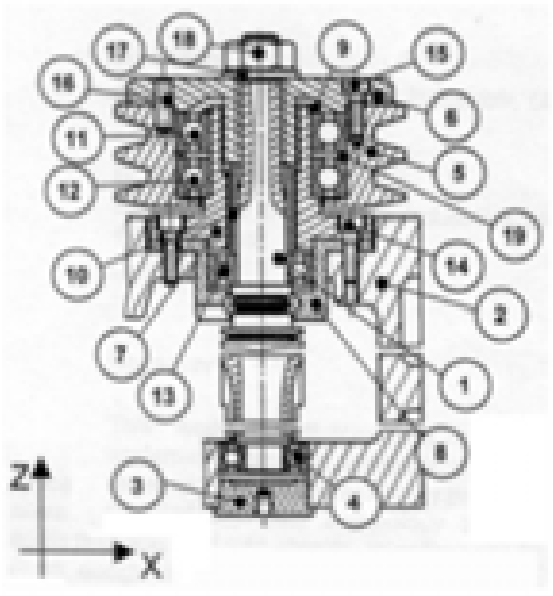

Fig. 4. Second Product

Some assumption were made to generate optimal assembly and disassembly sequences such as the product must be entirely disassembled (i.e. disassembly can be performed by reversing assembly sequence), the relationships between the components were known before the search of the optimal planning sequence, only technological and computation factors were considered in the fitness function (1). The proposed approach was developed starting from a well-known matrix-based GA methodology found in the literature (Dini G., et al. 1999). In this approach, a generic chromosome consisted of three sections that specify the component identifier, disassembly direction and selected gripper to be used for each part. The evaluation of the disassembly sequences was realized using the fitness function:

$$f = w_1 \cdot l + w_2 \cdot (N-1-o) + w_3 \cdot (N-g-1) \quad (2)$$

where constant N is the component number, while parameters l, o, g, s are, respectively, the maximum length of the feasible sequence, the orientation change number, gripper change number and maximum number of similar grouped disassembly components. The weights $w_i$ with i=1,3 are associated to each parameter.

Moreover an enhanced version with an adaptive fitness function found in (Lazzerini B. & Marcelloni F., 2000) was also considered. The difference with the previous fitness function was the creation of a function with a single term directly associated to the maximum length of the feasible sequence. In this way, individuals with a length less than the feasible one are removed from the initial population during evolution. Once the population contains all feasible individuals, the fitness function returns to the form (1). The advantage of using these twin functions was an improvement of GA convergence.

The Fuzzy-GA hybridization was implemented by the authors onto two levels.

The first level consisted of substituting the algebraic fitness function with a Fuzzy function to obtain closer control of the technological knowledge of the disassembly process. In fact normally there is not a unique definition of the weight affecting each technological factor (re-orientation, gripper changes, etc) and often these weights have to be changed depending on the plant layout and the availability of machines. Fuzzy classification systems based on fuzzy logic are capable of dealing with uncertainties and can be intuitively varied by an operator according to the machine availability. For each parameter, a fuzzy set with 3 triangular membership functions was created to specify the process condition (bad, medium or good).

The fuzzy sets of each parameter were then used as input of the Mamdani Inference system, obtaining a single de-fuzzified output, used to evaluate the quality of chromosomes generated by the GA.

As a consequence, the first advantage of this approach was the use of a non-weighted fitness function, avoiding solution dependency on the disassembly product characteristics because of the weight balancing. Moreover a better response was achieved thanks to the natural attitude of a Fuzzy Logic system to cope with multi-objective problems.

The second level of integration considered the fuzzy control of the genetic operators. As mentioned before, the algorithm was dramatically influenced by the imposed mutation probability and crossover rate. The developed fuzzy controller was able to modify these probabilities during algorithm execution. The Fuzzy Logic was embedded into the controller of a smart algorithm. The main features of this controller were the possibility of a better allocation of the algorithm resources to improve its performance, direct on-line tuning of Genetic Algorithm parameters and the comparison between off-line and on-line parameters for better control.

The main advantages offered by adapting GA parameters at run-time were the improved search in the solution space and a fast growth of the best chromosome fitness value independently of genetic parameters. The rules implemented on the fuzzy controller can be found in (Cordon et al. 1997). A GA-based software program was designed and developed in the MATLAB environment. The structure of the algorithm is represented in Fig. 5. Four choices were available:

A) Pure genetic algorithm with a single algebraic fitness function (2);

B) Genetic algorithm and a fuzzy ranking function;

C) Pure genetic algorithm with adaptive algebraic fitness function;

D) Adaptive fitness function and dynamic fuzzy-controlled genetic algorithm.

## 4. Discussion of results

The study was conducted neglecting the influence of the population size on the performance of the GAs. Population size normally has an influence on algorithm performances, but in this study the authors preferred to focus their attention on the aspects connected to the solution sensitivity due to mutation probability and crossover rate variations. As a consequence the initial population was set to 80 for all runs.

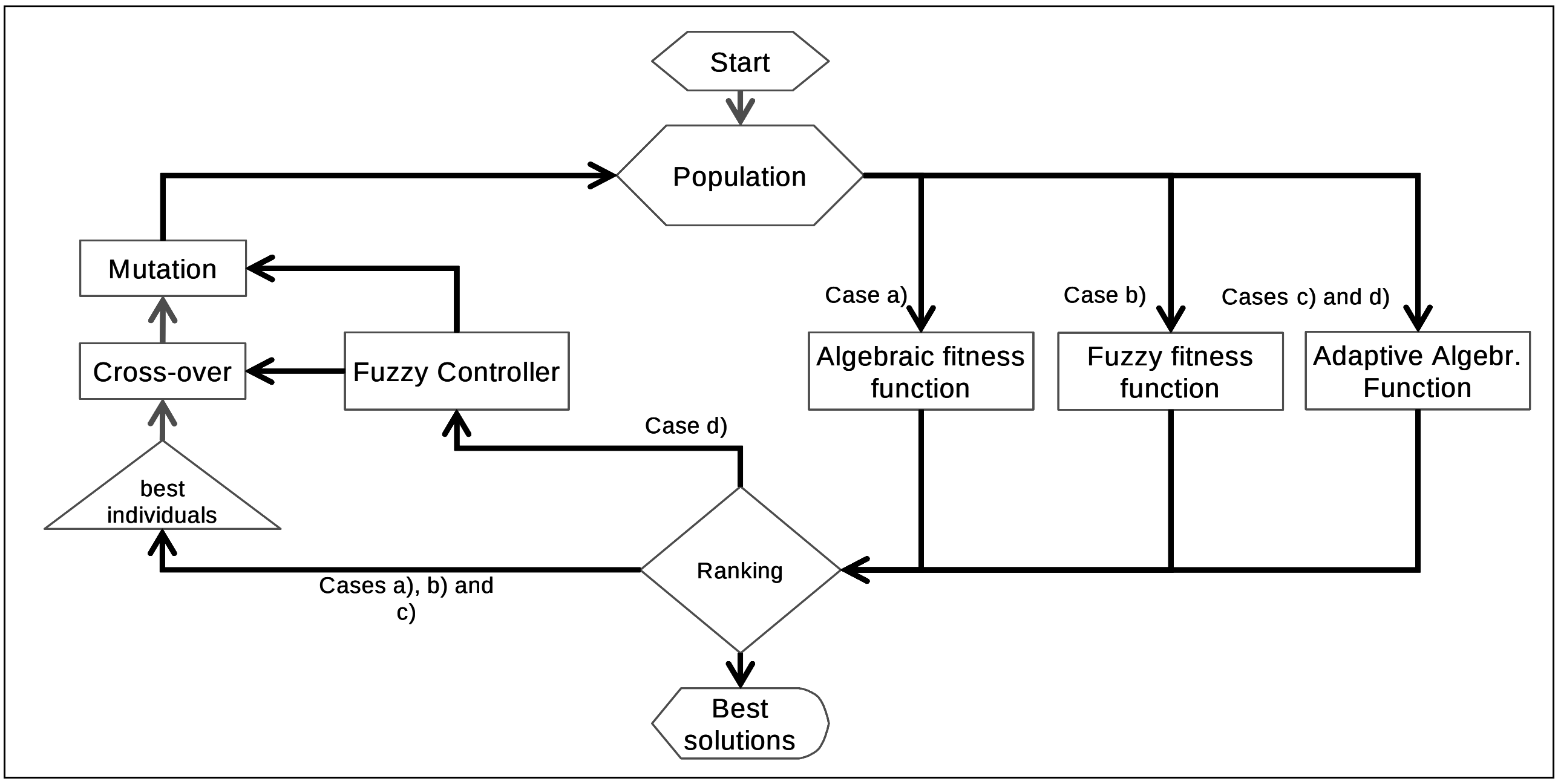

Fig. 5. Data flow diagram

*4.1 Choice A) versus Choice B)*

The pure genetic algorithm with the single algebraic fitness function (2) was compared to the genetic algorithm with a fuzzy ranking function (first level of hybridization). The comparison was carried out by evaluating results obtained from running both GA choices while keeping the same crossover rate, mutation probability and population size. In this way the same conditions were maintained for both investigated case studies. The parameters that led to best GA performance were used during the program running (mutation probability equal to 80% and a crossover rate equal to 40%). The runs were repeated 20 times each and the fitness function used to compare results was the mean of the maximum fitness for all the runs. The maximum fitness versus the generation for the first case study is showed in Fig. 6. Both the curves have similar behavior but the GA seems to behave slightly better than the Fuzzy. This behavior is confirmed by results in Fig. 7 where the maximum fitness versus the generation for the second case study is plotted. At first glance, the performance of the first level of integration seems to promote the use of pure GA. However these results must be carefully analyzed. In fact, a generic Fuzzy Function was employed while the parameters of the GA algebraic function were optimized for the assembly/disassembly applications.
The optimal tuning of GA parameters is not a simple operation but a very time-consuming task because results are only obtained after several GA runs have been performed. This task is applicable for products with a low number of components but becomes unfeasible, increasing of the component number.

Another aspect to consider is related to the number of successful identifications of the optimal sequence. Considering the second case study, the GA identified the optimal A/D sequence on 25% of the runs while GA with the Fuzzy function succeded for 60% of the runs.

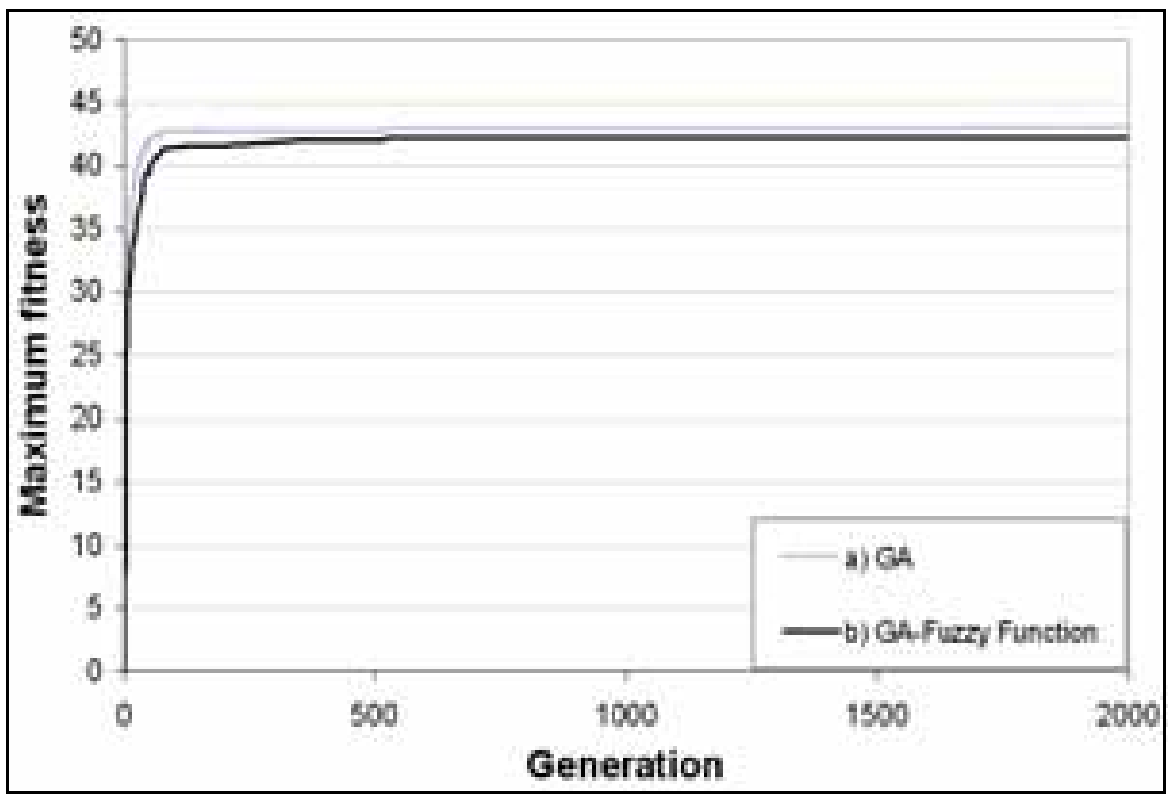

Fig. 6 Max. fitness vs generations, 1st case study

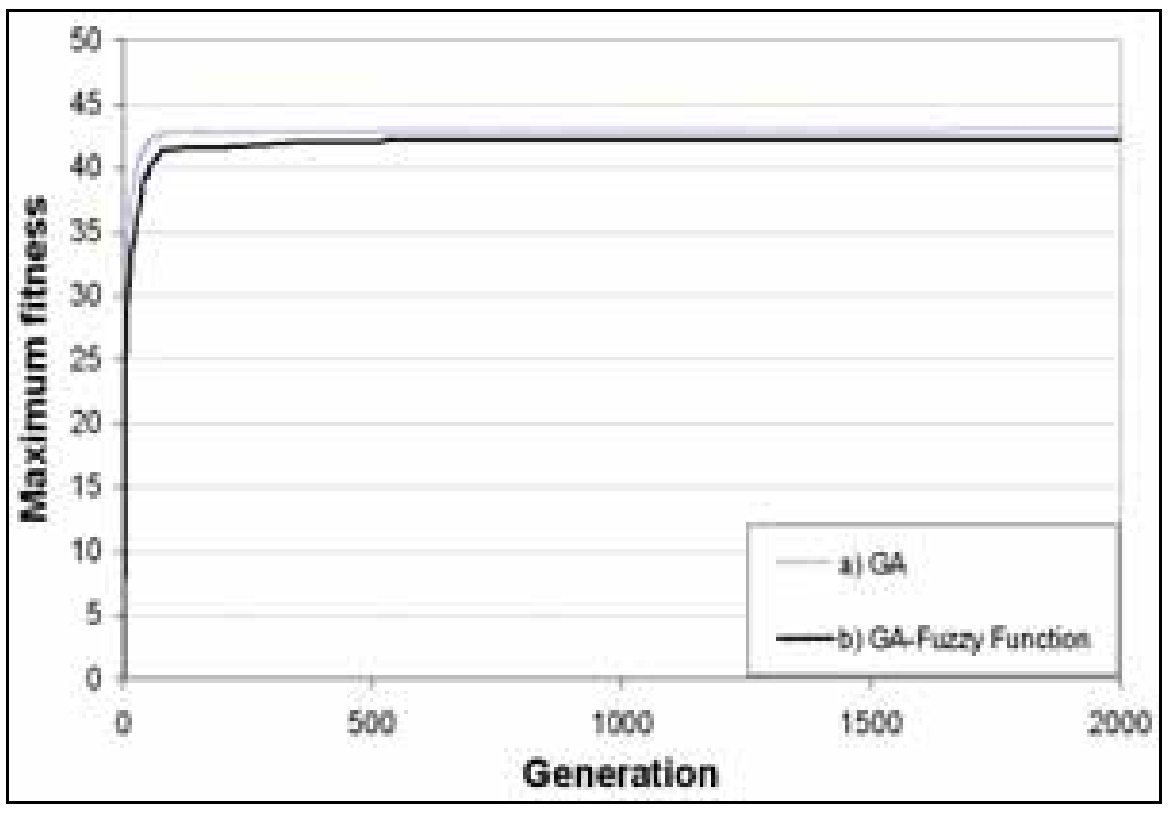

Fig. 7 Max. fitness vs generations, 2nd case study

The lower value of GA with Fuzzy function at the last generation can be explained by the fact that unfeasible sequences were obtained. The fitness of these sequences lowered the mean fitness function value used during comparison. On the contrary, the shape of the fitness function of GA eliminated unfeasible solutions and led to optimal or near-optimal solutions.

*4.2 Choice C) versus choice D)*

The performance of the Fuzzy controller was compared with the performances of GA with the adaptive fitness function. The comparison between these algorithms was performed by running algorithm C) 16 times (crossover rate 0.2-0.8 with step 0.2 and mutation probability 0.2-0.8 with step 0.2) for both case studies. The population size of the second case appears to be under-sized with respect to the first case. The results of the first and second case study are shown in Figure 8 and Figure 9 respectively. Different behaviour of the GA appears in both the situations.

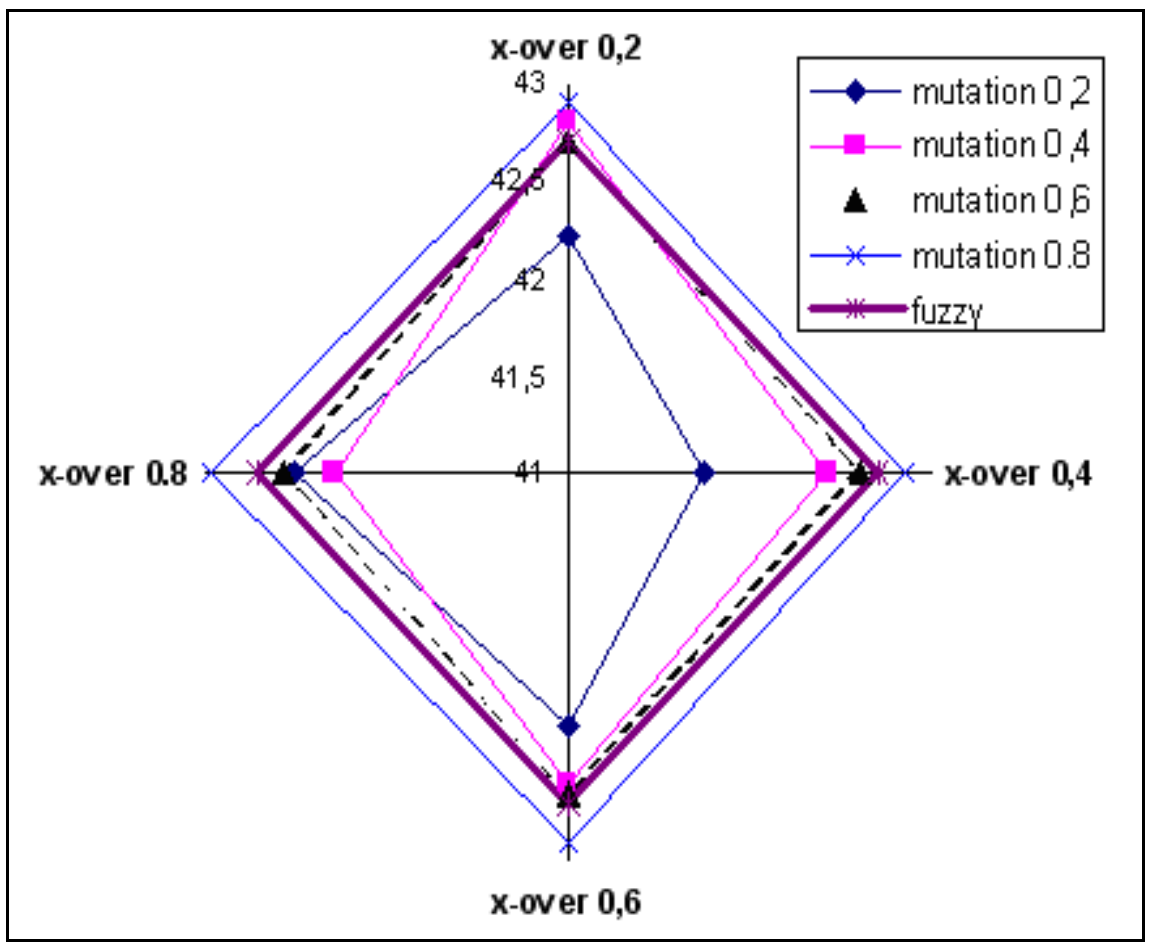

Fig. 8 Cross-over rate and mutation – 1st case study

During the identification of the optimal sequence of the first assembly the GA seems to behave well with a mutation probability equal to 0.8, independently of the value of the crossover rate.

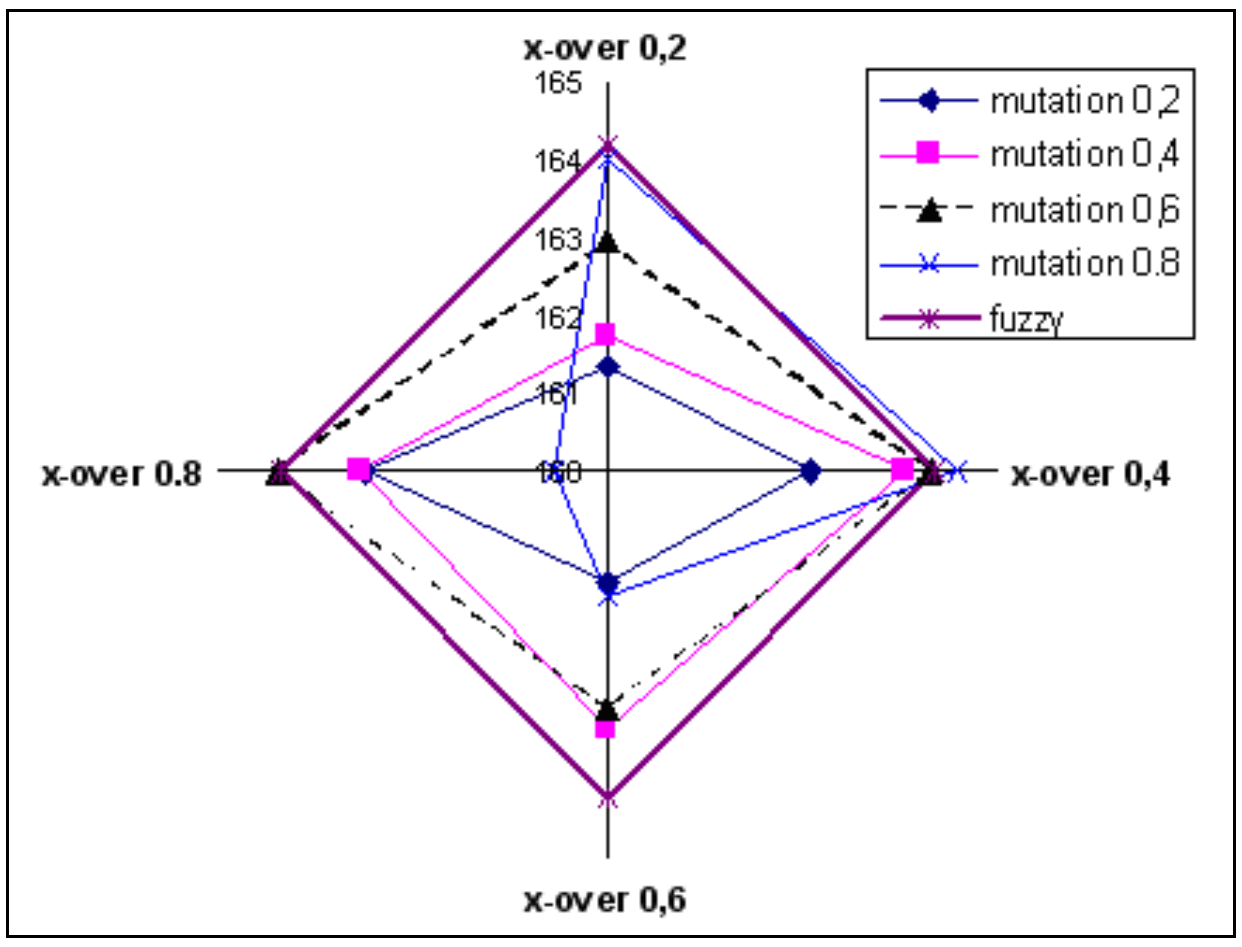

Fig.9: Cross-over rate and mutation – 2nd case study

The identification of the optimal sequence for the second case study was obtained with a mutation rate of 0.8 and a crossover rate of 0.4. The number of generations necessary to obtain the best results was equal to 5000 iterations. However, the results obtained using the Fuzzy-controlled GA proposed by the authors were better than those of pure GA and near to those obtained using the GA best performances in most cases. In Table 1 the optimal sequences found are shown for the two assemblies.

| First case study |
| --- |
| 8→7→5→6→4→2→3→1<br>+z→+z→+z→+z→+z→+z→+z→+z<br>G3→G1→G1→G1→G3→G1→G1→G1 |
| **Second assembly** |
| 3→4→16→15→18→17→6→10→9→11→<br>19→12→5→14→7→13→8→2<br>-z→-→z→z→z→z→z→z→z→z→z→z→z→<br>z→z→z→z→z→z<br>G5→G2→G2→G7→G8→G1→G1→G1→G6→G3→G3→G3→G3→G7→G3→G3→G3→<br>G3→G2 |

Table 1: Assembly sequences

## 5. Conclusions

Considering the results obtained using Fuzzy-GA, the stability and flexibility of the Fuzzy-GA approach for product assembly and disassembly planning can be further enhanced by studying a wider variety of product types and configurations. In particular the first level of hybridization gave similar results to the GA with the best parameter settings. On the other hand the second level denoted satisfying results, which can be enhanced with several improvements. The Fuzzy function can be improved by making it adaptive and more problem-oriented. The controller can also be improvable by making the population size and crossover probability adaptative as well.

## 6. References

Boothroyd G., Alting L., (1992), Design for assembly and disassembly, Annals of the CIRP, 41/2:625-636

Buker U. et alii (2001), Vision-based control of an autonomous disassembly station, Robotics and Autonomous Systems, Vol.35,pp179–189

Cordón O., Herrera F., Lozano M., (1997) On the Combination of Fuzzy Logic and Evolutionary Computation: A Short Review and Bibliography, In: Fuzzy Evolutionary Computation. Kluwer Academic Pub, pp. 33-56

Dini G., Santochi M., (1992) Automated sequencing and subassembly detection in assembly planning, Annals

of the CIRP, 41/1:1-4
Dini G., Failli F. (1998) A Computer Oriented Tool for Disassembly Oriented to Recycling of Products, Proceedings of CIRP-ICME 1998, pp. 599-606
Dini G., Failli F., Lazzerini B., Marcelloni F., (1999) Generation of optimized assembly sequences using genetic algorithms, Annals of the CIRP, 48/1:17-21
Dini G., Failli F. (1999) Searching Oprimized assembly Sequences by Ant Colony Systems, Proceedings of IV Aitem Conference, Brescia, Italy. pp.87-94,
Guide V.D.R. (2000), Production planning and control for remanufacturing: industry practice and research needs, J. of Operations Management, Vol.18,pp.467–483
Harjula T., Rapoza, B., Knight, W. A., Boothroyd, G., (1996) Design for Disassembly and the Environment, Annals of the CIRP Vol 45/1/1996, pp.109-114.
Herrera F., Lozano M., (1996) Adaptation of Genetic Algorithm Parameters Based on Fuzzy Logic, Genetic Algorithms and Soft Computing (F. Herrera, J. L. Vardegay (Eds.)), Springer-Verlag, pp. 95-125.
Kaebernick H., O'Shea B., Grewal S. (2000), A method for sequencing the disassembly of products, Annals of the CIRP,Vol.49/1,pp.13-15
Laperrière L., Elmaraghy H.A. (1992), Planning of products assembly and disassembly, Annals of the CIRP,Vol.41/1,pp.5-9
Lazzerini B., Marcelloni F., (2000) A Genetic Algorithm for generating Optimal Assembly Plans, Artificial Intelligence in Engineering, Vol. 14, pp. 319-329
Michalewicz Z. (1999), Genetic algorithm + data structures = evolution program, Spinger Verlag, ISBN 3-540-60676-9, USA
O'Shea B., Grewal S.S., H. Kaebernick (1998), State of the art literature survey on disassembly planning, Concurrent Engineering: Research and Applications, Vol.6, pp. 345-357.
Salomonski N., Zussman E. (1999), On-line predictive model for disassembly process planning adaptation, Robotics and Computer Integrated Manufacturing, Vol.15,pp.211-220
Santochi M., Dini G., Failli F. (2002), Disassembly for recycling, maintenance and remanufacturing: state of the art and perspectives, Proc. of 6th Int. Conf. on Advances Man. Systems and Tec. (AMST'02), Udine (Italy)
Srinivasan H., Figueroa R., Gadh R., (1999) Selective disassembly for virtual prototyping as applied to de-manufacturing, Robotics and Computer Integrated Manufacturing, Vol. 15, pp. 231-245.
Subramani A.K., Dewhurst P. (1991), Automatic generation of product disassembly sequences, Annals of the CIRP,Vol. 40/1,p.115-118